\documentclass[a4paper,fleqn]{cas-dc}

\usepackage[numbers]{natbib}
\usepackage{amsmath,amssymb,amsfonts}
\usepackage[ruled,vlined,linesnumbered]{algorithm2e}
\usepackage{subcaption}
\usepackage{tikz}
\usetikzlibrary{arrows.meta,positioning,shapes.geometric,backgrounds,fit,calc}
\usepackage{graphicx}
\usepackage{textcomp}
\usepackage{xcolor}
\usepackage{float}
\graphicspath{{./}}

\begin{document}
\let\WriteBookmarks\relax
\def\floatpagepagefraction{1}
\def\textpagefraction{.001}

\shorttitle{Lightweight ViT Compression for Plant Disease Detection}
\shortauthors{M.S Kumar et al.}

\title[mode=title]{Lightweight Vision Transformer Compression for On-Device Plant Disease Detection in Resource-Constrained Agricultural Field Conditions}

\tnotemark[1]
\tnotetext[1]{This work was supported by the ANRF, New Delhi, India, under Grant No.~SUR/2022/004268, ``Light Weight Deep Learning Based Mobile Application for the Early Detection, Identification, and Spatiotemporal Monitoring of Plant Diseases.''}

\author[1]{Mahadev~Sunil~Kumar}[orcid=0009-0005-4368-6537]
\ead{mahadevsunilkumar03@gmail.com}
\ead[url]{https://www.mahadevsunilkumar.com}
\credit{Methodology, Software, Writing -- Original Draft}

\author[1]{Bhavika~Gondi}
\ead{bhavikareddy.gondi@gmail.com}
\credit{Methodology, Software, Validation}

\author[1]{Desaisetty~Venkata~Satya~Sai~Swapnith}
\ead{swapnith07@gmail.com}
\credit{Methodology, Software}

\author[1]{Gangireddy~Rahul~Jogi}
\ead{rahuljogi756@gmail.com}
\credit{Methodology, Software}

\author[2]{Sudheesh~Manalil}
\ead{sudheeshmanalil@amrita.edu}
\credit{Resources, Data Curation, Funding Acquisition}

\author[3]{Arnab~Raha}[orcid=0000-0002-8848-1069]
\ead{arnab.raha@intel.com}
\ead[url]{https://sites.google.com/site/arnaverse}
\credit{Investigation, Writing -- Review \& Editing}

\author[4]{Amitava~Mukherjee}[orcid=0000-0003-1694-3911]
\ead{amitava@dubai.bits-pilani.ac.in}
\ead[url]{https://www.bits-pilani.ac.in/dubai/dr-amitava-mukherjee/}
\credit{Writing -- Review \& Editing, Supervision}

\author[5]{Parthasarathy~Seethapathy}[orcid=0000-0003-0660-9311]
\ead{parthasarathy.91@csir.res.in}
\credit{Resources, Writing -- Review \& Editing}

\author[1]{G.~Gopakumar}[orcid=0000-0003-1559-5150]
\cormark[1]
\ead{gopakumarg@am.amrita.edu}
\ead[url]{https://www.amrita.edu/faculty/gopakumarg/}
\credit{Conceptualization, Supervision, Project Administration}

\affiliation[1]{organization={Department of Computer Science and Engineering, Amrita School of Computing, Amrita Vishwa Vidyapeetham},
                city={Amritapuri},
                state={Kerala},
                country={India}}

\affiliation[2]{organization={Amrita School of Agricultural Sciences, Amrita Vishwa Vidyapeetham},
                city={Coimbatore},
                postcode={642109},
                state={Tamil Nadu},
                country={India}}

\affiliation[3]{organization={Intel Corporation},
                city={Santa Clara},
                state={California},
                country={USA}}

\affiliation[4]{organization={Department of Computer Science, Birla Institute of Technology and Science, Pilani Dubai Campus},
                city={Dubai},
                country={United Arab Emirates}}

\affiliation[5]{organization={Agrotechnology Division, CSIR--Institute of Himalayan Bioresource Technology},
                city={Palampur},
                postcode={176061},
                country={India}}

\cortext[cor1]{Corresponding author}

\begin{abstract}
	Chilli (\textit{Capsicum annuum}) is one of India's most economically significant crops, yet its productivity is persistently threatened by diseases that are difficult to identify without expert intervention. While Vision Transformers (ViTs) have achieved high classification accuracy, their large computational footprint makes deployment on resource constrained devices challenging. Existing compression approaches typically address pruning, quantization, and knowledge distillation in isolation, leaving the potential benefits and interactions of their combined application insufficiently explored. We propose a unified Vision Transformer compression framework that combines Hessian-Balanced Adaptive Block Pruning (H-BAC), guided by second-order sensitivity estimation, with quantization and attention-based knowledge distillation. To systematically identify the most effective configuration within each compression family, each technique is first evaluated independently through controlled ablation studies, after which the best-performing components are integrated into a sequential deployment pipeline tailored to real-world agricultural constraints. On a chilli 3-class village-split dataset with a genuine cross-village, cross-device out-of-distribution test split, the resulting compressed models match or exceed the 95.13\% FP32 baseline's accuracy, alongside 74--98\% model size reduction, and the fully integrated compression pipeline achieves a 54.5$\times$ size reduction (327.42\,MB to 6.01\,MB) at $95.13\pm2.32\%$ accuracy across four tested configurations. A direct comparison further reveals that, on this dataset, a directly-trained student of the same final size, without pruning or distillation, reaches comparable accuracy of 94.87\%, at the same 6.01\,MB INT8 size, indicating where H-BAC and knowledge distillation are, and are not yet shown to be, worth their computational cost.
\end{abstract}

\begin{highlights}
    \item H-BAC enables curvature-aware adaptive block pruning of Vision Transformers.
    \item H-BAC cuts ViT-B/16 FLOPs by 49\% while retaining 95.53\% OOD accuracy.
    \item The full pipeline compresses ViT-B/16 by 54.5$\times$ to a 6.01 MB model.
    \item Cross-village and cross-device testing evaluates real-world generalization.

\end{highlights}


\maketitle

\section{Introduction}
\label{introduction}

The major portion of India's economy is reliant on agriculture that is the main source of livelihood for nearly 55\% of the population. Among the crop varieties produced throughout the country, chilli (\textit{Capsicum annuum}) is of major importance because the country is the largest producer of chilli in the world and Andhra Pradesh on its own produces 49\% of the total amount at the national level~\cite{naik2022detection}. However, this crucial crop is constantly at the risk of diseases. Leaf Curl Virus, Anthracnose, and Bacterial Wilt are some of the diseases that result in a loss of 20\% to total crop failure depending on the severity of infection~\cite{naik2022detection}, and pest pressure compounds this risk further~\cite{wang2025advanced}. What makes this even more challenging is that many of these diseases look the same visually in the very early stages. Traditional diagnosing by manual expert checking or laboratory-based serological testing is very slow, far too costly, and highly dependent on having specialists available, which makes it not feasible at the scale of rural Indian farming~\cite{haque2025enhanced}.

To address this problem, it is necessary to have an automated system that can not only detect plant diseases but also identify them at the earliest stage in a direct manner in the field. The last ten years have witnessed tremendous strides in this area. At first, the use of traditional machine learning methods based on hand-crafted features like color histograms, texture descriptors and support vector machine classifiers were setting the stage but unfortunately, they did not generalize well to actual field conditions~\cite{shoaib2023advanced, ahmad2023survey}. It was the advent of Convolutional Neural Networks (CNNs) that transformed the scene: \cite{ferentinos2018deep} reached 95\% classification performance over 25 plant species, deeply rooting image-based deep learning as the prime approach. Further crop-specific designs even improved on this; for example, \cite{naik2022detection} reached 99.12\% on a chilli dataset. Vision Transformers (ViTs)~\cite{dosovitskiy2020image}, which model relationships across all image patches simultaneously through self-attention, have since pushed performance beyond 99\% on standard benchmarks~\cite{li2023pmvt, gole2023trincnet}, and now represent the state of the art for plant disease classification.

However, this state-of-the-art representational power comes at a steep computational cost. A standard ViT-B/16 occupies 327~MB and takes nearly 28~ms per CPU inference even on modern consumer hardware (Section~\ref{results}), making it incompatible with the low-end mobile devices that represent the primary computing resource available to most rural farmers~\cite{li2023pmvt, gole2023trincnet}. The gap between laboratory performance and field deployability is not a minor inconvenience; it is the central unsolved problem in applied agricultural AI.

Model compression techniques (pruning, quantization, and knowledge distillation) offer a good guide forward, but previous works see them as separated interventions~\cite{li2023model, kim2021pqk}. No one has, so far, conducted a joint evaluation of all three for ViT compression in agriculture where, besides variable lighting and class imbalance, device inference limitations add up to changes in what the general benchmark can depict~\cite{joshi2024edge}.

Although pruning, quantization, and knowledge distillation have been individually explored, existing works largely treat them as independent techniques and do not account for the varying sensitivity of transformer blocks. Moreover, conventional distillation approaches often suffer from feature mismatch when applied to heterogeneous architectures, and prior evaluations are typically limited to general-purpose benchmarks without considering real-world deployment constraints.

The main contributions of this work are as follows:
\begin{itemize}
\item We propose H-BAC, a curvature-aware adaptive block pruning method for Vision Transformers that combines Hutchinson-estimated block-level Hessian curvature with first-order Taylor-based component pruning within each block, and is, to our knowledge, the first such method evaluated as part of a joint compression pipeline for agricultural disease detection.
\item We introduce an attention-based knowledge distillation scheme that transfers relational reasoning through attention maps, bypassing the feature-dimension mismatch that causes conventional feature-based distillation to fail when teacher and student differ substantially in capacity.
\item We present the first joint evaluation of pruning, quantization aware training, and knowledge distillation for ViT compression under real-world agricultural constraints, providing a systematic comparison across all three techniques.
\end{itemize}

\section{Related work and research gaps}

\subsection{From traditional ML to vision transformers}

The detection of crop diseases using automated methods has evolved considerably over the past decade. Early approaches relied on traditional machine learning with hand-crafted features such as color histograms, texture descriptors, and SVM classifiers~\cite{cortes1995support,1017623,4309314}. Earlier plant disease studies combined classical machine learning with CNN-based feature learning, demonstrating that hybrid approaches can improve detection performance over hand-crafted feature methods~\cite{sajitha2022plant}. These methods, although feasible, were unable to generalize under real field conditions. Deep learning solved this problem by allowing an end-to-end feature learning that directly extracts features from images without the need for intermediate representation~\cite{shoaib2023advanced, ahmad2023survey}. Image-based deep learning frameworks have also been proposed for plant disease prediction, showing that end-to-end models can effectively learn discriminative disease features directly from leaf images~\cite{kirola2022plants}. \cite{ferentinos2018deep} found that CNNs could achieve 95\% accuracy in classifying 25 different plant species in the PlantVillage dataset~\cite{mohanty2016using}, establishing the convolutional approach as the reference standard in the field of agricultural image classification. Following that, domain-specific modifications: \cite{naik2022detection} through the integration of a Squeeze-and-Excitation CNN into a chilli dataset, reached an accuracy of 99.12\%, while \cite{hamim2024enhanced} developed a particular network architecture for chilli disease detection. Recent work has shown that CNN-based transfer learning can significantly improve plant disease classification accuracy~\cite{menon2021plant}. More recent crop-specific studies show that transformer-based and modern deep learning models are increasingly being adapted for agricultural diagnosis~\cite{megalingam2024cowpea}.

Nevertheless, CNN-based approaches are still limited by their local receptive fields, which is one of the reasons for changing the focus on attention-based architectures. Vision Transformers (ViTs), introduced by~\cite{dosovitskiy2020image}, overcome this limitation by considering an image as a set of patches where self-attention is applied to model local as well as global contexts. This characteristic is very advantageous for the task of plant disease detection, which requires relating the infection pattern to the overall leaf structure. \cite{singh2024effective} claimed that their GAN-augmented ViT could reach an accuracy of 99.92\% on PlantVillage, whereas \cite{barman2024vit} demonstrated tomato disease detection through a ViT-SmartAgri app on a smartphone at 90.99\% accuracy. \cite{haque2025enhanced} supplemented attention mechanisms to architectures used for multi-crop classification, beating previous best results on several benchmarks. \cite{hemalatha2024multitask} proposed a multitask ViT that jointly localizes and classifies plant disease regions, demonstrating that attention-based architectures can support diagnostic tasks beyond simple classification. However, standard ViTs require very high computational resources~\cite{thai2021artificial}. Crop-specific mobile disease diagnosis has also been explored for eggplant little leaf disease, reinforcing the need for deployment-aware model selection~\cite{seethapathy2025comparative}. \cite{gole2023trincnet} proposed TrIncNet, replacing Multi-Layer Perceptron (MLP) blocks with Inception modules to reduce model complexity. Newer framework-level studies continue to confirm the promise of deep learning for early detection of plant diseases, but most still do not directly address efficient deployment under strict edge-device constraints~\cite{vinay2025deep}.

\subsection{Lightweight architectures and model compression}

Plant-based Mobile ViT (PMVT)~\cite{li2023pmvt} fused the efficiency of MobileViT with Convolutional Block Attention Modules (CBAM), demonstrating acceptable accuracy-efficiency trade-offs. SLViT~\cite{li2023slvit} aimed at sugarcane leaf disease diagnosis, introduced shuffle operations to reduce computational complexity. A recent Edge-AI for agriculture study~\cite{joshi2024edge} showed that full-scale ViTs (323~MB) can reach 96\% accuracy but are unsuitable for resource limited settings, demonstrating that architectural efficiency alone is insufficient and post-training compression methods are required.

Model compression covers pruning, quantization, and knowledge distillation~\cite{li2023model}. The work by~\cite{han2015deep} on a combination of magnitude-based pruning with quantization and Huffman coding~\cite{huffman2024method} laid the groundwork, resulting in significant storage reductions with almost no accuracy loss. \cite{molchanov2019importance} further enhanced this with importance estimation based on Taylor expansion, including first-order gradient information for more principled pruning decisions. Closest to our pruning approach, \cite{yang2023global} (NViT) derives a Hessian-based structural saliency criterion for global ViT pruning with latency-aware regularization, redistributing parameters across blocks and within-block structures. H-BAC differs in three respects: it estimates block-level curvature via the Hutchinson trace estimator specifically to keep second-order estimation tractable at ViT-B/16 scale without a differentiable latency-regularization term, combines this with a separate first-order Taylor criterion for within-block component (attention head and MLP neuron) selection rather than a single unified saliency score, and is evaluated as one stage of a joint pruning-distillation-quantization deployment pipeline on a real-world agricultural edge-deployment task rather than as a standalone pruning method on general-purpose benchmarks. Knowledge distillation~\cite{hinton2015distilling} allows small student models to learn deeper inter-class relations from the teacher's soft probability outputs. \cite{paula2025comparative} revealed that the student can get closer to the teacher model's performance if temperature scaling and intermediate layer alignment are combined. Quantization-Aware Training (QAT) imitates low numerical precision during fine-tuning, so the model can update its weights before INT8 conversion~\cite{cheng2018model}. Each of these methods has been analyzed in isolation, meaning that whether their combination yields cumulative compression benefits remains an open research question.

Recent work demonstrates that combining compression techniques yields results beyond what any single method achieves. \cite{kim2021pqk} proposed PQK, a sequential pruning, quantization, and distillation pipeline. \cite{malihi2024matching, malihi2023efficient} showed that joint knowledge distillation and channel pruning consistently outperforms either applied individually. \cite{wang2025iterative} extended this to unsupervised domain adaptation through iterative KD and pruning under label scarcity. \cite{li2023constraint} demonstrated that OpenVINO's~\cite{intel2023openvino} Joint Pruning, Quantization, and Distillation toolkit achieves a 5.24$\times$ compression ratio with a 4.19$\times$ performance gain on BERT-base at under 1\% accuracy loss. In the context of agricultural disease detection using Vision Transformers, a joint evaluation of these combined compression methods remains unexplored. Despite these promising results, each approach treats pruning, quantization, and distillation as largely separable steps, optimized without regard for how one technique influences the behavior of the others. These interdependencies mean that sequentially applying well-performing individual techniques does not guarantee a well-performing joint solution, and can lead to compounding accuracy losses. Beyond this methodological limitation, none of the above works are evaluated under conditions representative of real agricultural deployment.

\subsection{Research gaps and our contributions}

Despite these advancements, there is still a major research gap in the systematic co-design of pruning, quantization, and distillation for ViTs in agricultural contexts. Most existing compression works are aimed at general-purpose benchmarks and overlook domain-specific challenges such as changing light conditions, uneven class distributions, and the need for explanation in farmer-facing tools~\cite{li2023model, joshi2024edge}. Furthermore, conventional ViT pruning techniques score first-order importance identically for all transformer blocks, thereby neglecting significant differences in block-level sensitivity. Conventional feature-based distillation similarly breaks down when teacher and student differ substantially in hidden dimension, as linear adapters cannot adequately bridge the feature-space mismatch.

This work addresses these gaps through a compression pipeline co-designed around ViT geometry rather than a sequential application of independent methods. The block-level curvature gap is addressed by H-BAC, which employs second-order Hessian curvature to assign non-uniform pruning rates across transformer blocks. The attention-based KD gap is targeted by an Attention-Based Knowledge Distillation scheme that transfers relational reasoning through multi-head attention maps, bypassing feature-dimension incompatibility entirely. Finally, the joint evaluation gap is filled by evaluating H-BAC, Attention-Based KD, and PTQ-Dynamic INT8 quantization together as a single deployment pipeline rather than as independent techniques.

\section{Methodology}
\label{sec:methodology}
\subsection{Dataset and preprocessing}

\begin{figure}[pos=h]
\centering
\includegraphics[width=0.8\columnwidth]{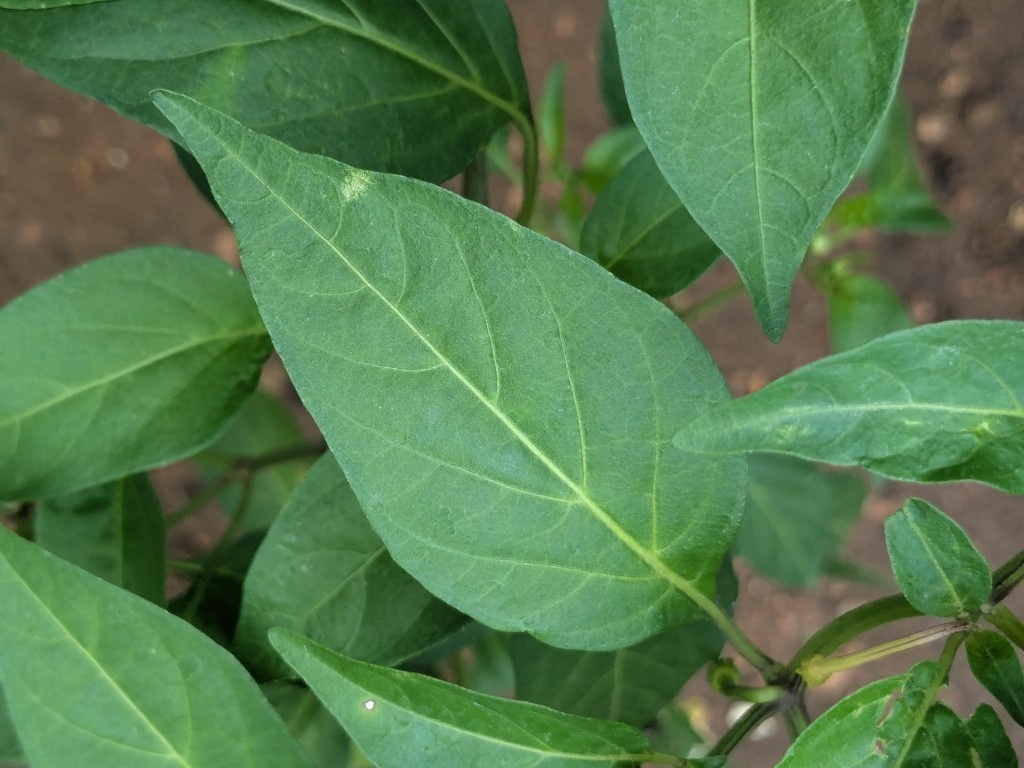}
\caption{Indigenously collected chilli leaf images captured in field conditions.}
\label{fig:custom_img}
\end{figure}

The dataset~\cite{kannan2026chilli} consists exclusively of locally captured images of diseased chilli leaves, taken on-site across smallholder farms in Coimbatore, Tamil Nadu, India, spanning three classes: Healthy control, Initial Symptoms of chilli leaf curl virus (ChiLCV), and Severe Symptoms of ChiLCV. These images reflect real-world acquisition conditions that include natural lighting variation, partial occlusion, complex backgrounds, and mixed disease symptom presentation: characteristics largely absent from controlled laboratory datasets. A selection of representative images from this collection is shown in Fig.~\ref{fig:custom_img}.

\begin{figure}[pos=h]
	\centering
	\includegraphics[width=\columnwidth]{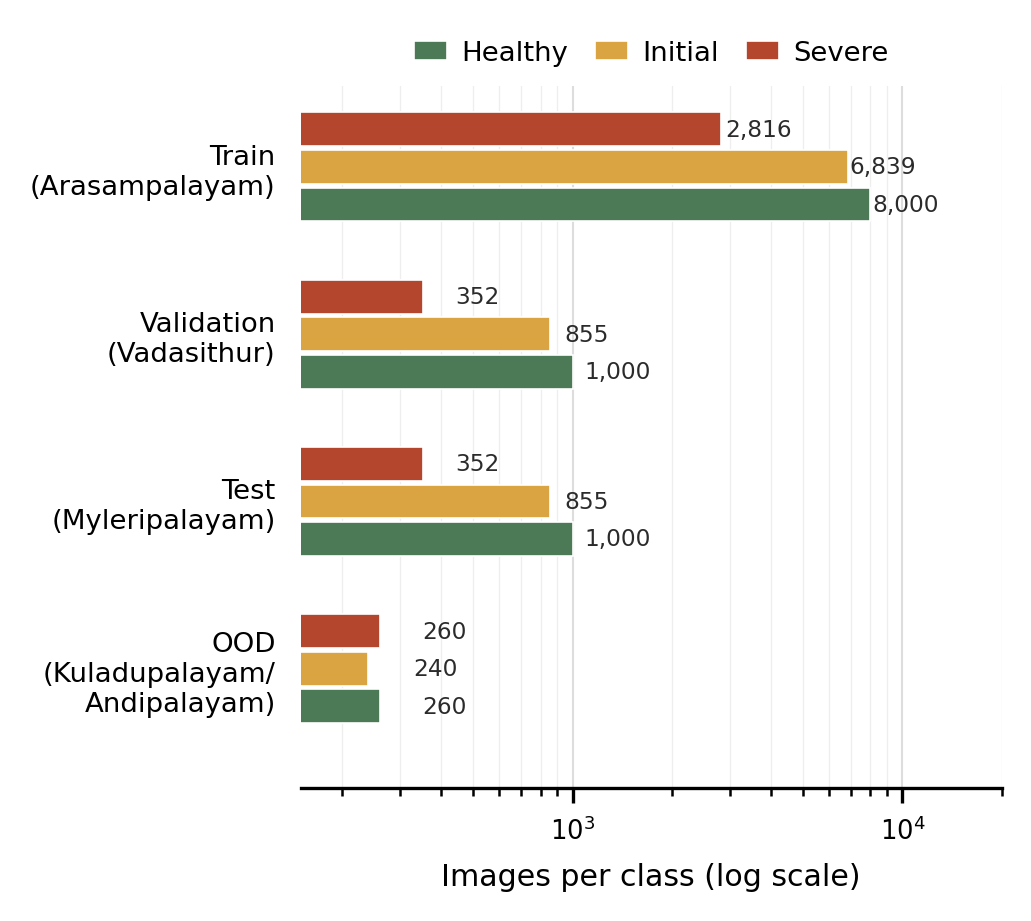}
	\caption{Per-class image counts across the four village-partitioned splits (log scale).}
	\label{fig:overall_dis}
\end{figure}

Unlike a pooled random split, every split in this dataset is drawn from a physically distinct village, and the Out of Distribution (OOD) split used for every reported result in this paper additionally spans different smartphone camera devices than training. The training split (17,655 images) was collected in Arasampalayam; the validation split (2,207 images) in Vadasithur; an in-distribution test split (2,207 images) in Myleripalayam; and a fourth, an out-of-distribution split (760 images) across two further villages, Kuladupalayam and Andipalayam, captured on three different phones (Realme C2, Redmi Note 12, Samsung Galaxy S23) not used to collect any other split. All four splits were collected between June and November 2024. Fig.~\ref{fig:overall_dis} visualizes the per-class composition of each split.

As every split is a physically distinct village, rather than a random stratified sample of a pooled collection, this dataset directly tests cross-domain generalization. Therefore, a model can no longer succeed by memorizing village-specific lighting, soil background, or camera-sensor characteristics that happen to recur across a random train and test portions. We exploit this directly by reporting all held-out accuracy, precision, recall, and F1 figures throughout this paper on the OOD split rather than the in-distribution Test split, since OOD additionally varies the acquisition device. This provides stricter and more deployment-realistic generalization test than an in-distribution village would provide on its own.

All images were resized to $224 \times 224$ pixels using bicubic interpolation to meet the patch embedding input requirements of the ViT-B/16 architecture~\cite{dosovitskiy2020image}. During training, random resized cropping with scale range (0.8, 1.0) and random horizontal flipping were applied to improve robustness to scale variation and viewpoint changes. Color jitter augmentation with brightness and contrast factors of 0.2 was applied to simulate natural lighting variation encountered in field settings. All images were normalized using ImageNet statistics~\cite{deng2009imagenet} (mean\,=\,[0.485, 0.456, 0.406], std\,=\,[0.229, 0.224, 0.225]).

\subsection{Baseline model}

ViT-B/16 was chosen as the high-accuracy baseline model pre-trained on ImageNet-1K~\cite{deng2009imagenet} due to its well-established performance on image classification tasks~\cite{dosovitskiy2020image}. To adapt the model for 3-class chilli disease classification, the final classification head was replaced with a fully connected layer with 3 output units, while all other pre-trained weights were retained and fine-tuned end-to-end on the Train split (Arasampalayam). The fine-tuned baseline, evaluated on the OOD split, achieved an accuracy of 95.13\%, a (weighted) precision of 95.55\%, recall of 95.13\%, and F1-score of 95.10\%. All latency figures in this paper use batched throughput timing: each timed call processes a batch of 128 images, and wall-clock time is divided by 128 to obtain a per-sample figure measurement. Reported values are the mean over independent runs (3 trials on CPU, 10 on GPU). Model training/fine-tuning and CPU inference latency measurement used an Apple M4 Pro (12-core CPU) MacBook Pro; GPU inference latency was measured separately on a NVIDIA RTX 4060 (8\,GB) instance. All CPU latency figures are measured through Apple's Core ML runtime (models converted with \texttt{coremltools}, executed on the CPU compute unit), which on this hardware is substantially faster than PyTorch's eager-mode CPU execution for every architecture tested here; GPU figures remain PyTorch/CUDA measurements, since Core ML targets only Apple's own compute units. The baseline performance and deployment characteristics are summarized in Table~\ref{tab:baseline}.

\begin{table}[pos=h]
\centering
\caption{Baseline ViT-B/16 Performance and Deployment Characteristics}
\label{tab:baseline}
\begin{tabular}{@{}ll@{}}
\hline
\textbf{Metric} & \textbf{Value} \\
\hline
OOD Accuracy      & 95.13\% \\
Precision (weighted) & 95.55\% \\
Recall (weighted)    & 95.13\% \\
F1-Score (weighted)  & 95.10\% \\
Model Size        & 327.42\,MB \\
Parameters        & 85.80M \\
CPU Latency (Apple M4 Pro, Core ML) & $7.18 \pm 0.01$\,ms \\
GPU Latency (NVIDIA RTX 4060) & $5.56 \pm 0.004$\,ms \\
\hline
\end{tabular}
\end{table}

The baseline fine-tuning recipe (AdamW, LR$=10^{-4}$, weight decay $10^{-4}$, label smoothing 0.1, cosine schedule with a 3-epoch warmup, early stopping on validation loss with patience 5) was used. We compared it, on the same held-out run, against four alternatives: a $3\times$ and $10\times$ lower learning rate ($3\times10^{-5}$ and $10^{-5}$), the lower learning rate combined with a dropout layer (0.1) before the classification head, and the lower learning rate combined with a $5\times$ higher weight decay ($5\times10^{-4}$). None of the four alternatives improved on the existing recipe (96.05\% OOD accuracy in this comparison).

Although the baseline achieves strong classification accuracy on genuinely unseen villages and devices, its large parameter count (85.80M) makes it a poor fit for the low-computation-power smartphones that most smallholder farmers in India rely upon. This motivates the compression pipeline described in the following subsections.

\begin{figure}[pos=h]
    \centering
    \includegraphics[width=\columnwidth]{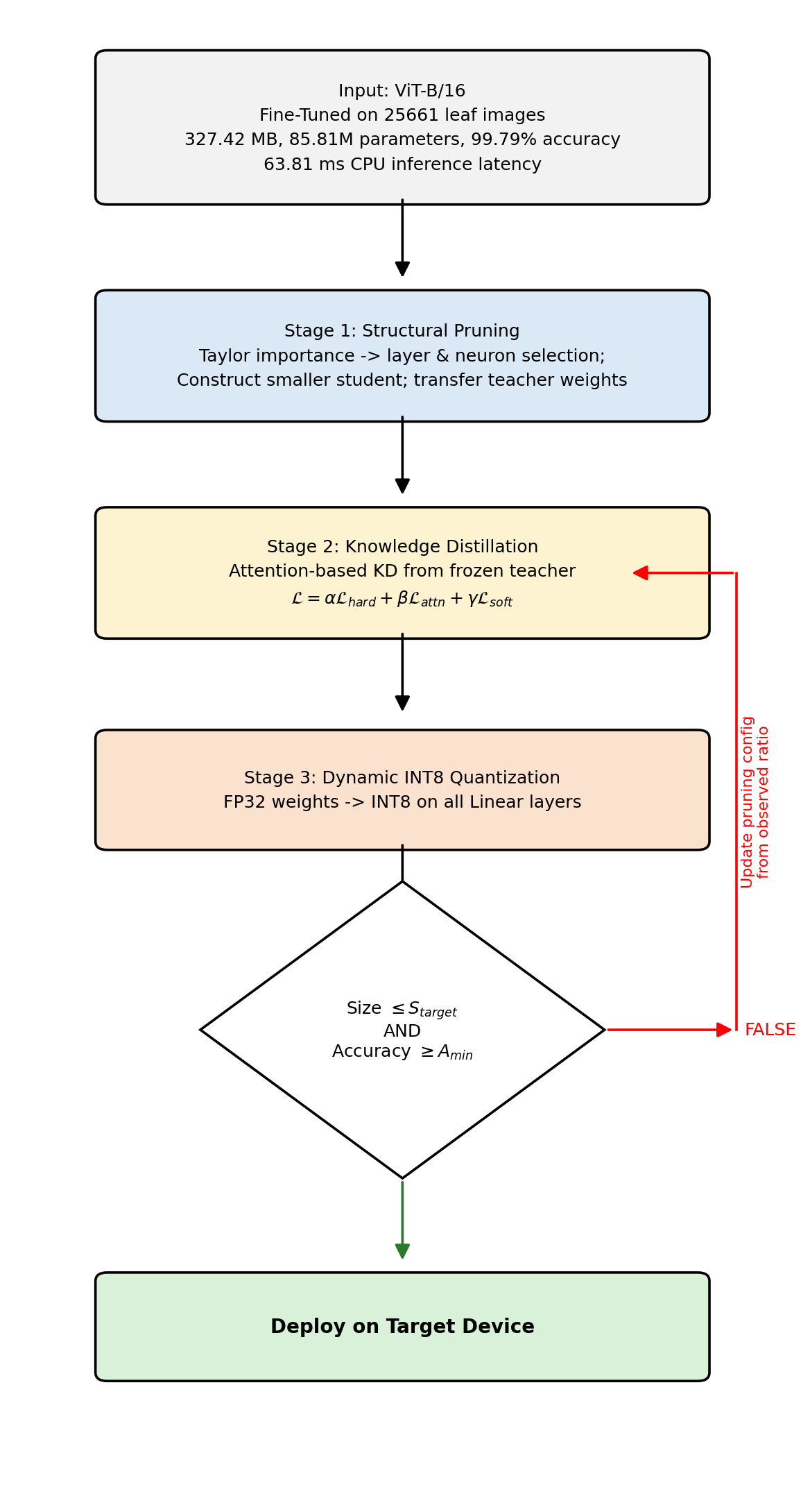}
    \caption{Workflow of a size/accuracy-budget-driven compression workflow: compress, check against target constraints, escalate if not met. Section~\ref{results} details the concrete 4-stage cascade used by the constrained search mode and its results.}
    \label{fig:pipeline}
\end{figure}

To handle the large model, we have applied three complementary model compression methods: Hessian-Balanced Adaptive Block Pruning (H-BAC), PTQ-Dynamic INT8 quantization, and Attention-Based Knowledge Distillation (KD). Fig.~\ref{fig:pipeline} illustrates the general principle of a size/accuracy-budget-driven deployment workflow: compress, check against the target size and minimum-accuracy constraints, and escalate to a more aggressive configuration if the constraints are not yet met. The paper reports two concrete uses of this principle. The primary one, evaluated throughout Section~\ref{results}, is the fixed Integrated Compression Pipeline described below (H-BAC at a single pre-selected pruning ratio, then KD, then PTQ-Dynamic quantization), which does not itself iterate. The second is a constrained search mode where we start from a size budget and an accuracy-drop budget and want the least aggressive configuration that meets both automatically; Section~\ref{results} reports this mode's concrete staged algorithm and grid-search results separately from the fixed pipeline's own numbers.

\subsection{H-BAC pruning}
\label{sec:hbac}

We present Hessian-Balanced Adaptive Block Pruning (H-BAC), the curvature-aware pruning method that reduces Vision Transformer FLOPs while preserving classification accuracy for edge deployment. Unlike prior work that applies uniform pruning across all transformer blocks, H-BAC incorporates second-order Hessian curvature information to allocate non-uniform, curvature-weighted pruning rates across blocks, and calculates importance using first-order Taylor series for component-level pruning within each block. The method proceeds through four phases: obtaining block-wise curvature via Hessian trace estimation, assigning pruning resources based on curvature, executing structural pruning on attention heads and MLP neurons, and performing fine-tuning for accuracy recovery.

Unlike first-order pruning methods that rely solely on gradient magnitudes, Hessian-based sensitivity measure captured through the Hessian matrix provides a more principled measure of parameter sensitivity. However, exact computation of the full Hessian is computationally intractable for large models such as ViT-B/16, as memory and compute requirements grow quadratically with the number of parameters~\cite{yang2023global}. To make second-order sensitivity estimation tractable, H-BAC employs the Hutchinson trace estimator~\cite{hutchinson1989stochastic}, which approximates the Hessian trace using random Rademacher vectors without explicitly forming the full Hessian matrix. This reduces the per-block curvature estimation cost from $\mathcal{O}(p^2)$ to $\mathcal{O}(Kp)$, where $p$ is the number of parameters in a block and $K$ is the number of Hutchinson samples. In practice, $K=10$ samples per block were found sufficient for stable curvature estimates.

We used the Hutchinson estimator~\cite{hutchinson1989stochastic} to approximate the curvature of each transformer block by calculating the trace of the Hessian:
\begin{equation}
    \mathrm{Tr}(H) \approx \frac{1}{K} \sum_{k=1}^{K} v_k^{T} H v_k
\end{equation}
where $v_k$ are random Rademacher vectors. For each transformer block, we compute the curvature separately for attention parameters and MLP parameters. The total block curvature is the weighted sum:
\begin{equation}
    C_b = H_{\text{attn}}^{b} + \lambda \cdot H_{\text{mlp}}^{b}
\end{equation}
where $\lambda$ balances the relative importance of the Multi-Layer Perceptron (MLP) component.
This coefficient is essential because the significant differences in parameter counts and gradient scales between the attention and MLP layers would otherwise cause the MLP block to dominate the overall sensitivity estimate.
To make rough second-order estimates more reliable, we take the absolute value of the curvature and average it over a small calibration dataset.

Unlike conventional pruning methods that apply uniform or magnitude-based criteria across all blocks, H-BAC uses block-level curvature as a sensitivity metric to determine how much each block should be pruned. Having curvatures of blocks $C_1, C_2, \ldots, C_B$, we normalise the pruning weights as $W_b = \sqrt{C_b}/\mathrm{mean\left(\sqrt{C}\right)}$.
Given a global pruning rate $r$, the pruning ratio $R_b$ assigned to block $b$ is: $R_b = r \cdot W_b$

Block ratios are clipped to $[0.05,\ 0.95]$, then rescaled so the mean ratio across blocks equals the global pruning target $r$, and finally re-clipped to the same bounds (clipping alone does not preserve the mean when curvature values are highly skewed across blocks, so an explicit renormalization step after clipping is required, and the rescaling can in turn push individual ratios back outside the bounds, so a second clip is applied; see Algorithm~\ref{alg:hbac}). Under this direct (non-inverted) weighting, higher-curvature blocks receive larger pruning ratios and lower-curvature blocks are pruned less. This was verified empirically. At a 50\% global ratio, pre-finetune accuracy under this direct weighting (64.87\%) exceeds a uniform-pruning baseline at the same ratio (41.32\%) by 23.55 percentage points. This direct-weighting formulation runs counter to the inverse-curvature intuition (protect high-curvature/high-sensitivity blocks) that motivates most Hessian-based pruning heuristics in the literature. A theoretically-motivated inverse-weighting variant, tested under identical conditions, performed worse on both ends (34.21\% pre-finetune, 94.21\% post-finetune) -- even underperforming uniform pruning pre-finetune -- confirming that direct weighting empirically identifies blocks that tolerate more pruning, not less.

At the component level, each transformer layer contains multiple attention heads that are rated using a first-order Taylor importance criterion~\cite{lecun1989optimal, molchanov2019importance}. For a particular attention head $h$, the importance score is:
\begin{equation}
I_j = \frac{1}{B} \sum_{b=1}^{B} \left\langle \frac{\partial \mathcal{L}}{\partial f_j^{(b)}}, f_j^{(b)} \right\rangle
\end{equation}
where \(f_j^{(b)}\) denotes the feature map produced by the \(j\)-th filter for the \(b\)-th sample in a mini-batch of size \(B\), \(\mathcal{L}\) is the classification loss, and \(\langle \cdot,\cdot \rangle\) denotes the Frobenius inner product. For block $b$, the number of heads pruned $k_b$ is:
\begin{equation}
    k_b = \left\lceil R_b \cdot H \right\rceil
\end{equation}
where $H$ is the total number of heads in the block. The least important heads are masked by zeroing the corresponding Query, Key and Value slices.

MLP neuron pruning is performed at the neuron level in the first fully connected layer of each transformer block. Neuron importance is estimated from a first-order Taylor approximation:
\begin{equation}
    I_n = \left| \frac{\partial \mathcal{L}}{\partial w_n} \odot w_n \right|
\end{equation}
where $w_n$ denotes the weights of neuron $n$. Neurons are ordered according to the L2 norm of their output weights, and those with the least importance are eliminated.

H-BAC uses mask-based structured pruning, where deactivated attention heads and MLP neurons are found by zeroing their weights while keeping tensor dimensions intact. During fine-tuning, the masks act as gradient masks to control the effect of pruning on training. An iterative short-retraining phase is conducted after the pruning steps to help the network recover from potential performance degradation. Fine-tuning is performed using a reduced learning rate with all remaining parameters unfrozen, and a single epoch was used in our testing.

\begin{algorithm}[t]
    \caption{H-BAC: Hessian-Balanced Adaptive Block Pruning}
    \label{alg:hbac}

    \KwIn{
        Pretrained ViT model $\mathcal{M}$ with $B$ transformer blocks\;
        Calibration dataset $\mathcal{D}_{calib}$\;
        Global pruning ratio $r$\;
        Hutchinson samples $K$\;
        MLP balance coefficient $\lambda$\;
        Per-block ratio bounds $\rho_{\min}=0.05$, $\rho_{\max}=0.95$
    }

    \KwOut{Pruned model $\mathcal{M}'$}

    Initialize block curvature vector
    $\mathbf{C} \leftarrow \mathbf{0} \in \mathbb{R}^{B}$\;

    \For{$b = 1$ \KwTo $B$}{

        $c_b \leftarrow 0$\;

        \ForEach{$(x,y) \in \mathcal{D}_{calib}$}{

            $\ell \leftarrow \mathcal{L}(\mathcal{M}(x), y)$\;

            $H_{\text{attn}} \leftarrow
            \mathrm{Tr}\!\left(
            \nabla_{\theta_{\text{attn}}^{b}}^{2}\ell
            \right)$ estimated using $K$ Hutchinson samples\;

            $H_{\text{mlp}} \leftarrow
            \mathrm{Tr}\!\left(
            \nabla_{\theta_{\text{mlp}}^{b}}^{2}\ell
            \right)$ estimated using $K$ Hutchinson samples\;

            $c_b \leftarrow
            c_b + H_{\text{attn}} + \lambda \cdot H_{\text{mlp}}$\;
        }

        $C_b \leftarrow \max(c_b,0)/|\mathcal{D}_{calib}|$\;
    }

    $\mathbf{W} \leftarrow
    \sqrt{\mathbf{C}}/\mathrm{mean}(\sqrt{\mathbf{C}})$\;

    $\mathbf{R} \leftarrow
    \mathrm{clip}(r \cdot \mathbf{W},\rho_{\min},\rho_{\max})$\;

    $\mathbf{R} \leftarrow
    r \cdot \mathbf{R}/\mathrm{mean}(\mathbf{R})$\;

    $\mathbf{R} \leftarrow
    \mathrm{clip}(\mathbf{R},\rho_{\min},\rho_{\max})$\;

    \For{$b = 1$ \KwTo $B$}{

        Compute first-order Taylor importance
        $I_i = \left| w_i \cdot
        \frac{\partial \ell}{\partial w_i} \right|$
        for attention heads in block $b$\;

        Prune $\lceil R_b \cdot H \rceil$
        least-important heads by zeroing corresponding
        QKV projection slices and output projection channels\;

        Compute first-order Taylor importance
        $I_i = \left| w_i \cdot
        \frac{\partial \ell}{\partial w_i} \right|$
        for MLP neurons in block $b$\;

        Prune $\lceil R_b \cdot N \rceil$
        least-important neurons by zeroing corresponding
        rows in \texttt{fc1} and columns in \texttt{fc2}\;
    }

    \Return $\mathcal{M}'$\;

\end{algorithm}
The hybrid design of H-BAC is crucial because it incorporates second-order curvature to reveal global sensitivity differences among the blocks, and uses first-order Taylor pruning as a lightweight mechanism for within-block pruning, producing higher pruning granularity and a better accuracy-efficiency trade-off. H-BAC exploits the curvature of the loss function to decide the pruning ratio of each transformer block, giving more pruning to high-curvature blocks and less to low-curvature, non-critical ones (see Section~\ref{sec:hbac} for the empirical validation of this direct weighting).

\subsection{Quantization strategies}
To further reduce memory usage and accelerate inference, two Post-Training Quantization (PTQ) variants were investigated: dynamic and static quantization, both applied directly to an already-trained FP32 checkpoint with no additional fine-tuning.

PTQ-Dynamic converts trained models to low precision without the need for retraining. Dynamic quantization was applied to the trained ViT-B/16 checkpoint, converting selected layer weights from FP32 to INT8 at inference time~\cite{paszke2019pytorch}. All Linear layers, including attention projections, MLP layers, and the classifier head were targeted, with weights quantized to QINT8 to reduce memory footprint and speed up linear operations. Activations remain in FP32, with scale factors computed dynamically at inference, requiring no calibration dataset.

PTQ-Static converts both weights and activations to INT8 before inference using a calibration dataset to compute fixed scale factors, offering greater speedup than dynamic quantization~\cite{paszke2019pytorch}. However, static quantization remains challenging for complex Vision Transformer architectures, primarily due to missing quantization operators for attention mechanisms and unsupported layer normalization operations. To address this, we propose a calibrated hybrid solution in which all linear layers are quantized to INT8, and a calibration dataset of 1,000 training images sampled using Weighted Random Sampling ensures class-balanced calibration across all 3 disease classes. We quantize every checkpoint in this study with PTQ-Dynamic applied directly to its final trained weights, with no extra fine-tuning stage, and no accuracy is given up once the checkpoint being quantized is already well-trained.

Full performance results across both quantization approaches are reported in Table~\ref{tab:quant_ablation} (Section~\ref{results}).

Between the two quantization candidates, PTQ-Dynamic (94.34\% accuracy, 84.42\,MB) and PTQ-Static (94.61\% accuracy, 85.78\,MB) both sit close to the FP32 baseline (95.13\%). PTQ-Static's 0.27-point edge is within normal run-to-run noise, and PTQ-Dynamic is what the rest of this pipeline (KD ablation, capacity sweep, constrained search) already uses consistently, so PTQ-Dynamic is adopted throughout for its simplicity. On CPU latency, the INT8-quantized model measures $7.21$\,ms (Apple M4 Pro via Core ML), showing no significant improvement over the FP32 baseline's $7.18$\,ms.

\subsection{Knowledge distillation}

Knowledge distillation (KD) transfers the knowledge of a large and powerful teacher model to a small student model. We evaluated three distillation approaches: response-based, feature-based, and attention-based, for compressing the ViT-B/16 teacher (85.80M parameters, 327.42\,MB) to a TinyViT student (5.52M parameters, 21.15\,MB), resulting in a 15.48$\times$ compression ratio. The student model \texttt{vit-tiny-patch16-224}~\cite{dosovitskiy2020image, winkawaks2022} was initialized from ImageNet-21k~\cite{5206848} pre-training with 5.52M parameters and 21.15\,MB.

\textbf{Response-Based KD:} Response-based KD uses the final output probability distributions (logits) for knowledge transfer without requiring internal model representations. For each training batch $t$, teacher logits $z^t \in \mathbb{R}^{3}$ are obtained by passing the input images through the frozen teacher model, and the same images are passed through the student model to obtain student logits $z^s \in \mathbb{R}^{3}$. Both sets of logits are divided by temperature $T=4.0$ before softmax for softer probability distributions:
\begin{equation}
p_i^t = \frac{\exp(z_i^t / T)}{\sum_{j=1}^{3} \exp(z_j^t / T)}, \quad p_i^s = \frac{\exp(z_i^s / T)}{\sum_{j=1}^{3} \exp(z_j^s / T)}
\end{equation}
Here, $z_i^t$ and $z_i^s$ denote the teacher and student logits, respectively, and $p_i^t$ and $p_i^s$ denote the corresponding softened probabilities. The combined loss balances learning from the ground truth and soft targets:
\begin{equation}
\mathcal{L}_{\text{RKD}} = \alpha \mathcal{L}_{\text{hard}} + (1-\alpha) \mathcal{L}_{\text{soft}}
\end{equation}
The hard loss $\mathcal{L}_{\text{hard}} = -\sum_{i=1}^{3} y_i \log(p_i^s)$ employs the true labels to guarantee correct prediction; the soft loss $\mathcal{L}_{\text{soft}} = T^2 \cdot \sum_{i=1}^{3} p_i^t \log\frac{p_i^t}{p_i^s}$ (Kullback-Leibler divergence) expresses to what extent the student imitates the teacher's distribution; and $\alpha=0.5$ weights both targets equally. $T^2$ scaling compensates for the gradient magnitude decrease that results from the temperature division.

\textbf{Feature-Based KD:} Feature-based KD targets the student model's learning at both the output level and the internal representation levels of the teacher model. Considering the hierarchical feature learning of Vision Transformers~\cite{raghu2021vit, dosovitskiy2020image}, we chose 4 intermediate transformer layers spaced at Layers 3, 6, 9, and 11 (out of 12). Since teacher features have 768 dimensions and the student has only 192 dimensions, adaptation layers are needed to bridge this gap, following practical guidelines established for ViT feature-based distillation~\cite{yang2022vitkd}. For each matched layer, we introduce a learnable adapter network consisting of a two-layer MLP with ReLU activation: Linear(192 to 768) followed by ReLU non-linearity, and Linear(768 to 768) for final projection. The combined feature KD loss $\mathcal{L}_{\text{FKD}}$ is:
\begin{equation}
    \mathcal{L}_{\text{FKD}} = \alpha \mathcal{L}_{\text{hard}}^{\text{FKD}} + \beta \mathcal{L}_{\text{feature}} + \gamma \mathcal{L}_{\text{soft}}
\end{equation}
where the feature loss $\mathcal{L}_{\text{feature}} = \frac{1}{4}\sum_{l \in \{3,6,9,11\}} \text{MSE}(\hat{f}_l^s, f_l^t)$ averages Mean Squared Error across 4 layers, and $\alpha=0.5$, $\beta=0.3$, $\gamma=0.2$, $T=4.0$.

\textbf{Attention-Based KD.} Unlike prior feature-based approaches, attention-based KD works by transferring knowledge via the distribution of attention weights rather than raw feature values. The attention weights reveal which patches interact strongly (e.g., diseased region attending to healthy regions for contrast), enabling the student to learn the teacher's visual reasoning process. For each transformer layer, self-attention computes:
\begin{equation}
\text{Attention}(Q, K, V) = \text{softmax}\!\left(\frac{QK^T}{\sqrt{d_k}}\right)V
\end{equation}
The attention matrix $A = \text{softmax}(QK^T/\sqrt{d_k}) \in \mathbb{R}^{197 \times 197}$ represents pairwise token relationships. Since the teacher has 12 attention heads per layer and the student has only 3, the attention weights are averaged across all heads within each layer:
\begin{equation}
\bar{A}^t = \frac{1}{12}\sum_{h=1}^{12} A_h^t, \quad \bar{A}^s = \frac{1}{3}\sum_{h=1}^{3} A_h^s
\end{equation}
For each matched layer $l \in \{3, 6, 9, 11\}$, teacher and student multi-head attention maps are collected and averaged over heads to produce $\bar{A}_l^t$ and $\bar{A}_l^s$. The combined attention KD loss $\mathcal{L}_{\text{AKD}}$ is:
\begin{equation}
    \mathcal{L}_{\text{AKD}} = \alpha \mathcal{L}_{\text{hard}}^{\text{AKD}} + \beta \mathcal{L}_{\text{attention}} + \gamma \mathcal{L}_{\text{soft}}
\end{equation}
where the attention loss $\mathcal{L}_{\text{attention}} = \frac{1}{4}\sum_{l \in \{3,6,9,11\}}$ and $\text{MSE}(\bar{A}_l^s, \bar{A}_l^t)$ averages Mean Squared Error between attention maps, and $\alpha=0.5$, $\beta=0.3$, $\gamma=0.2$, $T=4.0$.

Full performance results across all knowledge distillation approaches are reported in Table~\ref{tab:kd_ablation_combined} (Section~\ref{results}).

\begin{table}[pos=h]
    \centering
    \caption{H-BAC 50\% Pruning -- Classification Performance (post-finetune)}
    \label{tab:hbac_50_perf}
    \begin{tabular}{@{}lcc@{}}
        \hline
        \textbf{Model} & \textbf{Accuracy (\%)} & \textbf{FLOPs} \\
        \hline
        ViT-B/16 Baseline        & 95.13      & 35.13\,GFLOPs \\
        ViT-B/16 + H-BAC (50\%)  & 95.53 & 17.86\,GFLOPs \\
        \hline
    \end{tabular}
\end{table}

\begin{table*}
    \caption{H-BAC 50\% Pruning -- Deployment Characteristics, uncompacted mask-based model (CPU: Apple M4 Pro via Core ML; GPU: NVIDIA RTX 4060 via PyTorch)}
    \label{tab:hbac_50_deploy}
    \centering
    \begin{tabular}{lcccc}
        \hline
        \textbf{Model} & \textbf{Active Params} & \textbf{Size (MB)} &
        \textbf{CPU (ms)} & \textbf{GPU (ms)} \\
        \hline
        ViT-B/16 Baseline       & 85.80M     & 327.42 & 7.18    & 5.560 \\
        ViT-B/16 + H-BAC (50\%) & 43.70M & 327.42 & 7.18  & 5.560 \\
        \hline
    \end{tabular}

    \vspace{0.5ex}
    {\footnotesize Latency here is measured on the mask-based model \emph{before} structural compaction (pruned rows/columns still physically present, zeroed), matching the pipeline stages of Table~\ref{tab:integrated_pipeline_stages}. The two rows are, by construction, the same architecture (mask-based pruning zeroes weights without changing tensor shapes), so they get an identical measured latency here -- this is deliberately not the deployment-facing number, which Table~\ref{tab:hbac_ablation} reports after structural compaction.}
\end{table*}

\subsection{Integrated Compression Pipeline}

The proposed compression framework follows a two-stage optimization strategy designed to ensure principled component selection before pipeline integration.
Initially, the three compression families -- pruning, quantization, and knowledge distillation, are evaluated independently through controlled ablation experiments. This step determines the maximum performance achievable and side-effects related to the deployment for each individual technique, which allows for a fair comparison of compression families without the influence of confounding interactions.

PTQ-Dynamic is used for the pipeline's final quantization stage. An extra fine-tuning pass before quantization only matters on checkpoints that have not yet had adequate training, and every checkpoint reaching this stage of the pipeline already has KD's 15 distillation epochs. PTQ-Dynamic is applied directly with no added training step. Attention-Based KD is selected among distillation strategies for its capacity to transfer relational reasoning.
In the second stage, the selected components are composed sequentially to construct the final deployment-ready model. The full-precision ViT-B/16 teacher undergoes H-BAC structural pruning to reduce its computational footprint while preserving important transformer blocks. Attention-Based Knowledge Distillation is subsequently applied, imparting the reasoning patterns of the frozen pruned teacher to the student model. Finally, PTQ-Dynamic quantization is applied directly to the distilled student to obtain an INT8-precision model suitable for deployment on resource-constrained hardware. This sequential composition ensures that after each stage, the working representation is further compressed and that the handling of compression technique interactions is done in a way that is controlled and reproducible rather than the joint optimization disregarding their interdependencies.

Table~\ref{tab:integrated_pipeline_stages} reports the stage-by-stage outcome of this pipeline at the 50\% H-BAC operating point, measured on the OOD split throughout. Starting from the pipeline's own 95.13\% / 327.42\,MB FP32 entry point, H-BAC pruning followed by one epoch of recovery fine-tuning yields 94.34\% accuracy at unchanged mask-based size (43.70M of 85.80M parameters active). Attention-Based KD distills this into a 21.15\,MB TinyViT student at 93.68\% accuracy, and the final PTQ-Dynamic quantization step, applied directly with no further fine-tuning, produces a 6.01\,MB deployed model at 91.97\% accuracy, corresponding to a 54.5$\times$ size reduction relative to the FP32 checkpoint size.

Repeating the full pipeline across three additional configurations yields accuracies of 91.97\% (50\% pruning, Table~\ref{tab:integrated_pipeline_stages}), 97.24\% and 96.45\% (two further 50\% runs), and 94.87\% (70\% pruning). Across the four tested configurations, the final-stage accuracy is therefore $95.13\pm2.32\%$. The traced 50\%-pruning configuration gives the lowest accuracy at 91.97\%, suggesting some variability in final-stage performance across runs.

%Repeating this full pipeline across three additional configurations (pruning ratio 70\% at seed 42; seeds 123 and 456 at 50\%) gives a final-stage accuracy of $95.13\pm2.32\%$ (mean $\pm$ std, $n=4$);the seed-42 run above is the weakest of the four (91.97\%), while seeds 123 and 456 reach 97.24\% and 96.45\% respectively and the 70\%-ratio seed-42 stress config reaches 94.87\%, indicating meaningful run-to-run variance at this final stage, consistent with the seed sensitivity also observed in the capacity-tier student sweep (Section~\ref{results}).

We additionally tested the simplest alternative to this pipeline, skipping H-BAC pruning and KD entirely, by training a TinyViT student directly on labels and subsequently quantizing it. As shown in the final row of Table~\ref{tab:integrated_pipeline_stages}, this direct-training alternative reaches $94.87\%$ accuracy at the same 6.01\,MB model size. This is comparable to the $95.13\pm2.32\%$ accuracy obtained across the four pipeline configurations, while avoiding the additional pruning and distillation stages. Section~\ref{results} discusses this comparison and its implications in detail.

\begin{table*}[h]
    \caption{Integrated Compression Pipeline: stage-by-stage results (H-BAC 50\%, OOD split; CPU: Apple M4 Pro via Core ML; GPU: NVIDIA RTX 4060 via PyTorch)}
    \label{tab:integrated_pipeline_stages}
    \centering
    \begin{tabular}{lcccccc}
        \hline
        \textbf{Stage} & \textbf{Accuracy (\%)} & \textbf{Size (MB)} & \textbf{Active Params (M)} & \textbf{CPU (ms)} & \textbf{GPU (ms)} \\
        \hline
        FP32 Baseline$^{\dagger}$        & 95.13 & 327.42 & 85.80 & 7.18 & 5.560 \\
        H-BAC Pruned (pre-finetune)      & 64.87 & 327.42 & 43.70 & 7.18 & 5.560 \\
        H-BAC Pruned (finetuned)         & 95.53 & 327.42 & 43.70 & 7.18 & 5.560 \\
        + Attention-Based KD             & 93.68 & 21.15  & 5.52  & 1.02 & 0.660 \\
        + PTQ-Dynamic (deployed)         & 91.97 & 6.01   & 5.52  & 1.02 & --   \\
        \hline
        \emph{Direct-training alternative}$^{*}$ & \emph{94.87} & \emph{6.01} & \emph{5.52} & \emph{1.02} & --- \\
        \hline
    \end{tabular}

    \vspace{2pt}
    \footnotesize
    $^{*}$Skips H-BAC pruning and KD entirely: TinyViT trained directly on ground-truth labels,
    then quantized with PTQ-Dynamic, applied directly with no extra fine-tuning.
    $^{\dagger}$Measured through the pipeline's own data loader, which applies the same
    background-suppression preprocessing used to originally train this checkpoint -- see the
    baseline-model note earlier in this section. The pre-finetune row is measured directly from
    the saved intermediate checkpoint.
    All latency figures use batched throughput timing (batch of 128, wall time divided by 128)
    to amortize fixed per-call dispatch overhead; CPU and GPU latency are pending
    re-measurement on the target hardware (see baseline-model note).
\end{table*}

\section{Results}
\label{results}

\subsection{Experimental setup}

All experiments are carried out with the ViT-B/16 backbone fine-tuned on the chilli 3-class village-split dataset described in Fig.~\ref{fig:overall_dis} (17,655 training images; all reported accuracy/precision/recall/F1 figures below are measured on the 760-image OOD split). H-BAC serves as a post-training, mask-based structured pruning method. Global pruning ratios from 10\% to 90\% with a step of 10\% were evaluated; ratios beyond 90\% were not swept in this study. Model training/fine-tuning and CPU inference latency measurement used an Apple M4 Pro (12-core CPU) MacBook Pro; GPU inference latency was measured separately on a NVIDIA RTX 4060 (8\,GB). These measurements characterize each compression technique's relative efficiency gain (size reduction, FLOPs reduction, and cross-method latency ranking) under controlled, reproducible conditions, not the absolute latency a farmer's device would see. Validating the deployed model on actual ARM smartphone hardware is a scope boundary of the present study, discussed further in Section~\ref{results}.

\begin{figure}[pos=h]
    \centering
    \includegraphics[width=\linewidth]{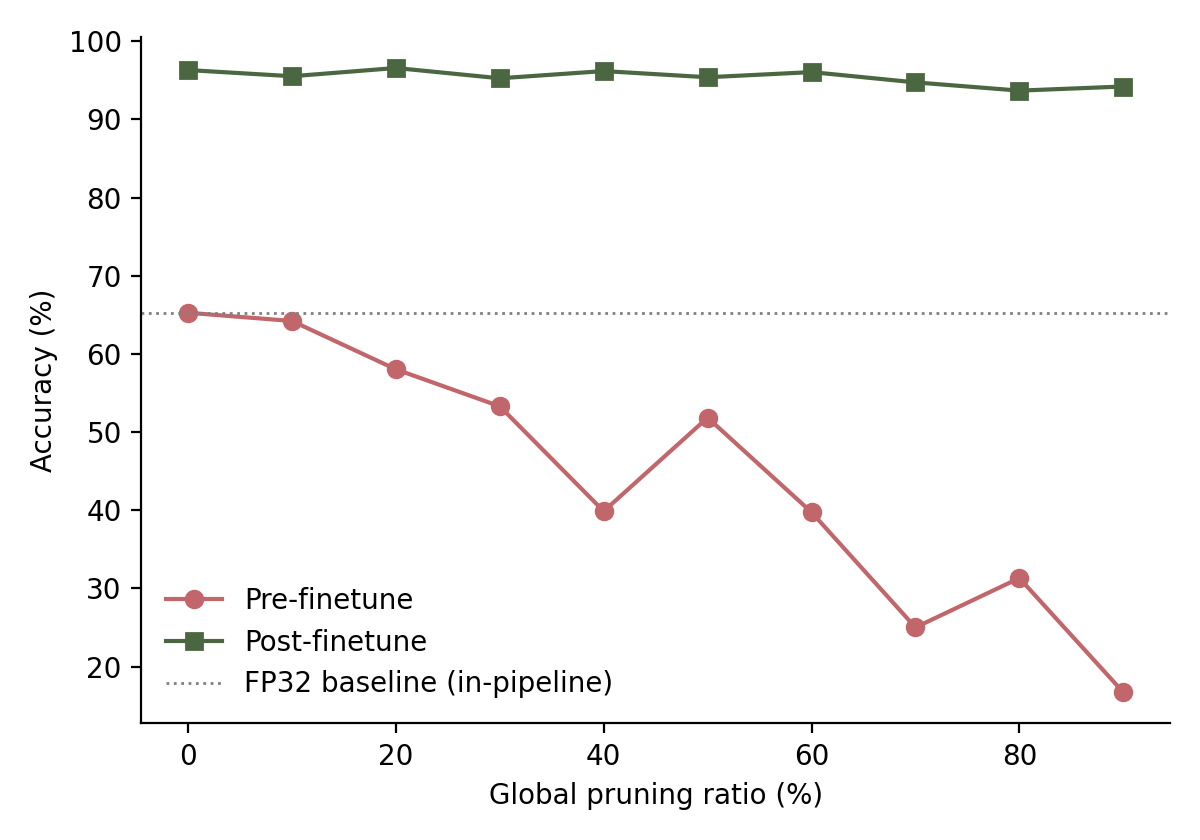}
    \caption{OOD accuracy (\%) of H-BAC pruned ViT-B/16 at each global pruning ratio, before and after one epoch of recovery fine-tuning. Both curves are measured.}
    \label{fig:Accuracy}
\end{figure}

Fig.~\ref{fig:Accuracy} shows the pre-/post-finetune accuracy trends discussed below. Because H-BAC's mask-based pruning zeroes weights without shrinking the underlying tensors, latency was measured on a structurally compacted copy of each pruned model, consistent with the deployment-facing sizes reported below. Table~\ref{tab:hbac_ablation} reports both the pre-finetune and post-finetune accuracy at each global pruning ratio, since H-BAC's recovery fine-tuning stage substantially changes the accuracy-ratio trade-off. All accuracy figures are measured on the OOD split.

\begin{table*}
    \caption{Ablation of H-BAC pruning ratio. FLOPs and active parameters (mask-based model) decrease with pruning ratio; CPU/GPU latency measured on a structurally compacted copy (CPU: Apple M4 Pro via Core ML, 3 trials; GPU: NVIDIA RTX 4060 via PyTorch, 10 trials).}
    \label{tab:hbac_ablation}
    \renewcommand{\arraystretch}{1.15}
    \setlength{\tabcolsep}{4pt}
    \centering
    \small
    \begin{tabular}{ccccccc}
        \hline
        \textbf{Pruning ratio (\%)} & \textbf{Active params (M)} &
        \textbf{FLOPs (G)} &
        \textbf{Acc.\ pre-FT (\%)} & \textbf{Acc.\ post-FT (\%)} & \textbf{CPU (ms)} & \textbf{GPU (ms)} \\
        \hline
        0 (baseline) & 85.80 & 35.13 & 95.13 & 93.95 & 7.13 & 5.560 \\
        10           & 77.14 & 31.57 & 94.87 & 96.45 & 6.76 & 5.034 \\
        20           & 68.56 & 28.04 & 89.34 & 96.32 & 6.13 & 4.554 \\
        30           & 60.34 & 24.68 & 73.03 & 94.61 & 5.50 & 4.027 \\
        40           & 51.72 & 21.13 & 68.68 & 97.37 & 4.88 & 3.494 \\
        \textbf{50}  & \textbf{43.70} & \textbf{17.86} &
        \textbf{64.87} & \textbf{95.53} & \textbf{4.27} & \textbf{3.003} \\
        60           & 36.30 & 14.81 & 53.42 & 94.08 & 3.62 & 2.489 \\
        70           & 29.37 & 11.98 & 34.21 & 91.97 & 3.27 & 2.138 \\
        80           & 25.75 & 10.50 & 33.03 & 93.68 & 2.96 & 1.888 \\
        90           & 22.88 & 9.32  & 34.21 & 91.84 & 2.62 & 1.649 \\
        \hline
    \end{tabular}

    \vspace{0.5ex}
    {\footnotesize Sparsity (\%) is omitted as it is directly derivable from active params: $1 - \text{active}/85.80$. All latency figures use batched throughput timing (batch of 128, wall time divided by 128).}
\end{table*}

Pre-finetune accuracy declines steeply and steadily with pruning ratio, from 95.13\% at 0\% to a near-random floor of 33--34\% by 80--90\% pruning. This is the expected signature of removing function before any recovery step is applied: heavier pruning strips away more of the network's learned capacity, and pre-finetune accuracy tracks that loss almost monotonically across the sweep.

Post-finetune, the picture changes entirely: accuracy is remarkably stable across the full ratio sweep, ranging only 91.84--97.37\% with no visible degradation trend even at 90\% pruning (91.84\%, within 2 points of the least-aggressive configurations tested). One epoch of recovery fine-tuning is sufficient to erase nearly all of the pre-finetune ratio-dependence on this dataset. In practice, this gives a practitioner considerable headroom: once recovery fine-tuning is part of the pipeline (as it always is in the integrated pipeline of Table~\ref{tab:integrated_pipeline_stages}), the pruning ratio's dominant effect is on FLOPs and active-parameter count rather than on final accuracy, so the ratio can be pushed well past this study's chosen 50\% operating point -- trading further FLOPs reduction for essentially no measured accuracy cost.

H-BAC's curvature-aware allocation was compared against a uniform-pruning baseline at the same 50\% global ratio (identical first-order Taylor ranking within each block, but every block assigned the same pruning rate rather than H-BAC's curvature-weighted allocation). H-BAC outperforms uniform pruning at both stages of the pipeline: pre-finetune, 64.87\% vs.\ 41.32\% (a 23.55-point margin) at near-identical active parameter counts (43.70M vs.\ 43.30M); post-finetune, 95.53\% vs.\ 93.16\% (a 2.37-point margin). Curvature-guided budget allocation therefore protects accuracy more effectively than a flat allocation both before and after recovery training, confirming that the block-level sensitivity information H-BAC estimates via the Hutchinson trace is worth the extra computation it costs.

\subsection{Quantization results}

Two quantization strategies were evaluated against the FP32 baseline: dynamic post-training quantization (PTQ-Dynamic) and calibrated static post-training quantization (PTQ-Static). Both target INT8 precision on all Linear layers and are applied directly to the trained FP32 checkpoint, with no additional fine-tuning step. Table~\ref{tab:quant_ablation} summarizes the results.

\begin{table*}[h]
    \caption{Ablation of quantization methods (CPU: Apple M4 Pro via Core ML, 3 trials). Both methods
    achieve a $\sim$74\% model size reduction; the difference is in accuracy retention.}
    \label{tab:quant_ablation}
    \renewcommand{\arraystretch}{1.2}
    \centering
    \setlength{\tabcolsep}{4pt}
    \small
    \begin{tabular}{lccccc}
        \hline
        \textbf{Method} & \textbf{Size (MB)} &
        \textbf{Acc. (\%)} & \textbf{$\Delta$Acc. (\%)} &
        \textbf{Latency (ms)} & \textbf{Speedup} \\
        \hline
        FP32 Baseline     & 327.42 & 95.13 & ---     & 7.18 & 1.00$\times$ \\
        \textbf{PTQ-Dynamic}       & \textbf{84.42}  & \textbf{94.34} & $\mathbf{-}$\textbf{0.79} & \textbf{7.21}$^{a}$ & \textbf{1.00$\times$} \\
        PTQ-Static        & 85.78  & 94.61 & $-$0.52 & 7.21$^{a}$ & 1.00$\times$ \\
        \hline
    \end{tabular}

    \vspace{0.5ex}
    {\footnotesize $^{a}$Core ML applies a single weight-only INT8 scheme, so both PTQ variants
    convert to the same INT8 weight representation and share one measured latency; the accuracy
    figures are from the respective PyTorch PTQ implementations, which do differ.}
\end{table*}

Both methods produce a $\sim$74\% model size reduction, as the change in weight precision is the primary driver of compression, not structural modifications (PTQ-Static's slightly larger footprint reflects the small overhead of its embedded quantization observers/scale metadata). Each costs a small amount of accuracy relative to the FP32 baseline (PTQ-Dynamic $-$0.79 points, PTQ-Static $-$0.52 points) -- the expected signature of converting an already-fixed set of FP32 weights to INT8 with no further adaptation. PTQ-Static's 0.27-point edge over PTQ-Dynamic is within normal run-to-run noise; we adopt PTQ-Dynamic throughout this study (Table~\ref{tab:integrated_pipeline_stages} onward) for its simplicity, since it requires no calibration dataset.

On latency, INT8 quantization shows no significant improvement over FP32 inference on this hardware ($1.00\times$; $7.21$\,ms vs.\ $7.18$\,ms). Because these figures use batched throughput timing (batch of 128, wall time divided by 128), fixed per-call dispatch overhead is already amortized out of the measurement, so this reflects the compute cost of Core ML's weight-only INT8 path, which dequantizes weights back to floating point for the matrix multiply rather than executing native INT8 arithmetic, leaving per-operation cost essentially unchanged. This is an important consideration for practitioners targeting ARM-based edge hardware: on this platform, INT8 quantization's benefit is almost entirely in \emph{model size} (74\% reduction, critical for storage- and bandwidth-constrained devices), not inference speed.

\subsection{Knowledge distillation results}

Three distillation strategies, plus a no-distillation control, were evaluated for compressing the ViT-B/16 teacher (327.42\,MB, 85.80\,M parameters, 95.13\% OOD accuracy) to a TinyViT student (21.15\,MB, 5.52\,M parameters): response-based KD, feature-based KD, and attention-based KD. All configurations achieve an identical 15.48$\times$ compression ratio. Table~\ref{tab:kd_ablation_combined} reports mean $\pm$ standard deviation over 3 independent runs per method.

\begin{table*}[h]
    \centering
    \caption{KD Ablation Study: Model Size, Performance, and Speedup Comparison (CPU: Apple M4 Pro via Core ML, 3 trials), mean $\pm$ std over 3 runs}
    \label{tab:kd_ablation_combined}
    \renewcommand{\arraystretch}{1.2}
        \begin{tabular}{@{}lllll@{}}
            \hline
            \textbf{Method} & \textbf{Size (MB)} & \textbf{Compression Ratio} & \textbf{Acc. (\%)} & \textbf{CPU Speedup$^{b}$} \\
            \hline
            Teacher (ViT-B/16)    & 327.42 & $1.00\times$  & $95.13$ & $1.00\times$  \\
            TinyViT (no KD)$^{a}$ & 21.15  & $15.48\times$ & $96.58 \pm 0.26$ & 7.07$\times$ \\
            Response KD  & 21.15  & $15.48\times$ & $96.27 \pm 0.65$ & 7.07$\times$ \\
            Feature KD            & 21.15  & $15.48\times$ & $96.18 \pm 0.60$ & $7.07\times$ \\
            \textbf{Attention KD} & \textbf{21.15} & \textbf{$15.48\times$} & \textbf{$96.71 \pm 1.03$} & \textbf{$7.07\times$} \\
            \hline
        \end{tabular}%

    \vspace{0.5ex}
    {\footnotesize $^{a}$ Trained using ground truth labels only. $^{b}$ Speedup computed from batched-throughput CPU inference latency (batch of 128, wall time divided by 128); the teacher's CPU latency in this benchmark run was 7.18\,ms, TinyViT's 1.02\,ms. All four student configurations share the identical TinyViT architecture, so CPU speedup is identical across rows by construction; only the learned weights (and therefore accuracy) differ.}
\end{table*}

All four training recipes cluster tightly, within a 0.53-point band of each other (96.18--96.71\%), with Attention-Based KD achieving the highest mean accuracy of the four ($96.71\pm1.03\%$), narrowly ahead of direct training on ground-truth labels alone ($96.58\pm0.26\%$, no KD), response KD ($96.27\pm0.65\%$), and feature KD ($96.18\pm0.60\%$). Attention-Based KD's advantage over the no-KD control is modest (0.13 points) but consistent with attention transfer being the more effective of the three distillation signals tested here: it is also the only KD variant with a higher mean than the no-KD control, while response KD and feature KD both trail it. Attention-Based KD does carry the widest run-to-run spread of the four (std 1.03 vs.\ 0.26 for no-KD), meaning its advantage should be read as a consistent edge across configurations rather than a guaranteed per-run win.

Beyond the standalone comparison above, the question that matters for deployment is whether Attention-Based KD earns its place inside the full integrated pipeline, where it distills from an H-BAC-pruned-and-finetuned teacher rather than the full unpruned one, and is followed directly by PTQ-Dynamic quantization. Training a fresh TinyViT student directly on ground-truth labels and quantizing it with the same PTQ-Dynamic step used by the pipeline's final stage yields a 6.01\,MB model at \textbf{94.87\% accuracy} (Table~\ref{tab:integrated_pipeline_stages}'s direct-training alternative row). Compared against the matched single pipeline run (91.97\%), this direct-training alternative is 2.89 points ahead; compared against the mean of four independent pipeline configurations ($95.13\pm2.32\%$, three at a 50\% pruning ratio and one at 70\%), it is instead 0.26 points \emph{behind}. The full pipeline's run-to-run variance is therefore the more important effect here than any fixed accuracy cost or benefit from chaining H-BAC and Attention-Based KD ahead of quantization: averaged over multiple runs, the integrated pipeline is modestly ahead of simply training the target-sized architecture directly and quantizing it, even though any single pipeline run can land above or below that simpler alternative.

\begin{table}[h]
    \centering
    \caption{Capacity sweep: Attention-Based KD vs.\ no-KD direct training (both taken through PTQ-Dynamic quantization, applied directly with no extra fine-tuning), four from-scratch (non-pretrained) student sizes below TinyViT's 5.52\,M parameters, mean $\pm$ std over 3 runs}
    \label{tab:capacity_sweep}
    \renewcommand{\arraystretch}{1.2}
    \begin{tabular}{@{}lccc@{}}
        \hline
        \textbf{Params} & \textbf{No-KD (\%)} & \textbf{Attn-KD (\%)} & \textbf{Advantage (pp)} \\
        \hline
        1.31\,M & $77.24\pm13.75$ & $70.26\pm19.15$ & $-6.97\pm32.59$ \\
        1.89\,M & $70.57\pm19.35$ & $76.80\pm15.86$ & $+6.23\pm34.25$ \\
        2.63\,M & $77.15\pm6.01$ & $81.32\pm8.92$ & $+4.17\pm5.41$ \\
        \textbf{3.16\,M} & \textbf{$90.35\pm0.88$} & \textbf{$82.11\pm8.07$} & \textbf{$-8.25\pm8.06$} \\
        \hline
    \end{tabular}
\end{table}

To test whether Attention-Based KD's advantage over direct training grows as student capacity shrinks below TinyViT's 5.52\,M parameters, we swept four smaller student configurations (1.31\,M to 3.16\,M parameters), each trained from scratch (no ImageNet-pretrained checkpoint exists for these custom architectures, so both the no-KD and Attention-KD runs at a given size start from identical random initialization, a fair like-for-like comparison) and quantized directly with PTQ-Dynamic, across 3 runs per size. Table~\ref{tab:capacity_sweep} reports the resulting means. Run-to-run variance dominates this sweep: at three of the four sizes (1.31\,M, 1.89\,M, 2.63\,M) A fairly consistent effect emerges is in the largest tested, 3.16\,M parameters, where Attention-KD trails direct training by a margin close to its own noise ($-8.25\pm8.06$ points across runs). Training these small, non-pretrained architectures from scratch is evidently a much less stable process than fine-tuning an ImageNet-pretrained backbone. Deploying at these sizes should budget for multiple training runs rather than trusting a single run, and should not expect a clean, monotonic capacity-dependent trend from this recipe on this dataset.

Attention-Based KD from the pruned-and-finetuned teacher reaches 93.68\%, 1.84 percentage points \emph{below} Attention-Based KD run from the unpruned teacher under otherwise identical settings (95.53\%). This suggests that, at least on this dataset, whatever generalization capacity is lost through H-BAC pruning and its single-epoch recovery fine-tune is not fully passed on for the student to recover during distillation. Quantizing the KD student directly with PTQ-Dynamic reaches 91.97\%, matches or slightly exceeds a fine-tune-then-quantize variant we tested on the same checkpoint (91.45\%). This is consistent with the pattern established in \ref{sec:methodology}. The extra fine-tuning pass is only useful on checkpoints that have not yet had adequate training, and the KD student, which already distilled for 15 epochs, does not need.

\subsection{Constrained search results}
\label{sec:constrained-search}

Beyond the fixed pipeline evaluated above, we implemented a constrained search mode for the scenario where a a maximum deployment size and a maximum acceptable accuracy drop from the FP32 baseline is specified. The least aggressive configuration that satisfies both automatically, rather than committing to the full H-BAC+KD+quantization pipeline by default is needed. The search cascades through candidate families of increasing aggressiveness, stopping at the first one that satisfies both constraints: (1) H-BAC structural pruning alone (FP32), (2) H-BAC pruning plus dynamic INT8 quantization, (3) the KD-distilled TinyViT student (FP32), (4) the KD-distilled student plus INT8 quantization, and, should none of these four satisfy the target, four additional from-scratch student architectures below TinyViT's capacity (the 3.16\,M, 2.63\,M, 1.89\,M, and 1.31\,M-parameter students of Table~\ref{tab:capacity_sweep}, largest first), each tried in both FP32 and INT8 form -- twelve candidate stages in total. Within families 1 and 2, where a continuous pruning ratio must be chosen, we exploit the fact that both size and accuracy are monotonically non-increasing in pruning ratio (Table~\ref{tab:hbac_ablation}): the ratio that just barely satisfies the size budget is therefore also the ratio with the best accuracy among every ratio that fits, so a bisection search over the ratio finds the optimal configuration within a family in $O(\log(1/\epsilon))$ evaluations rather than a linear sweep. Each capacity-tier stage trains its candidate once per target with a freshly drawn random initialization rather than the fixed initializations used elsewhere in this study, since this mode answers "can a feasible model be found" as a live search rather than reproducing a specific prior run.

We evaluated 5 size targets (10, 15, 20, 25, 30\,MB) $\times$ 2 accuracy-drop budgets (2\% and 3\%), 10 combinations in total. All 10 were satisfied, but the tightest and loosest budgets in this grid are met by  different mechanisms. At the two loosest size budgets (25 and 30\,MB), the cascade stops at stage 3: the plain FP32 KD-distilled student (21.15\,MB, 93.68\% accuracy, a 1.45-point drop from baseline) already fits, so no quantization is needed. At the three tightest budgets (10, 15, and 20\,MB), stage 4 -- the same KD student, quantized to INT8 -- is not enough: at 6.01\,MB it comfortably fits the size budget, but its 3.16-point accuracy drop exceeds even the looser 3\% target, so the search escalates further. Stages 5 through 8 (the two largest capacity-tier students, FP32 and INT8) are each tried and rejected on accuracy, not size: all four reach their size targets easily but sit 8.7--11.8 points below baseline. The search succeeds at stage 9, the 1.89\,M-parameter capacity-tier student (FP32, no-KD strategy): 7.27\,MB at 94.08\% accuracy, only a 1.05-point drop -- comfortably inside every budget from 10\,MB upward. Table~\ref{tab:constrained_search} summarizes the grid; Table~\ref{tab:search_trace} traces stages 3 through 9 of the escalation for the tightest target (10\,MB, 2\% drop). No target in this grid was satisfied at stage 1 or 2 (H-BAC pruning alone, with or without quantization): H-BAC's mask-based pruning does not reduce stored model size at all, so a pruned-but-unquantized ViT-B/16 stays at 327.42\,MB regardless of ratio, far above every budget tested here.

\begin{table*}[h]
    \centering
    \caption{Constrained search: full 10-target grid (5 size budgets $\times$ 2 accuracy-drop budgets). All 10 combinations were satisfied.}
    \label{tab:constrained_search}
    \renewcommand{\arraystretch}{1.2}
    \begin{tabular}{@{}lcccc@{}}
        \hline
        \textbf{Size budget} & \textbf{Stage reached} & \textbf{Family} & \textbf{Acc.\ (\%)} & \textbf{Size (MB)} \\
        \hline
        10--20\,MB  & 9 & 1.89M-param student (FP32, no-KD) & 94.08 & 7.27 \\
        25--30\,MB  & 3 & KD student (FP32)  & 93.68 & 21.15 \\
        \hline
    \end{tabular}
\end{table*}

\begin{table}[h]
	\centering
	\caption{Escalation trace for the tightest target in Table~\ref{tab:constrained_search} (10\,MB, 2\% accuracy-drop budget), stages 3 through 9. Stages 1--2 (H-BAC pruning, with or without quantization) already fail every budget in this grid on size alone and are omitted; stages 5--8 fit the size budget but fail the accuracy budget.}
	\label{tab:search_trace}
	\renewcommand{\arraystretch}{1.2}
	\small
	\resizebox{\columnwidth}{!}{%
		\begin{tabular}{@{}clccc@{}}
			\hline
			\textbf{Stage} & \textbf{Family} & \textbf{Acc.\ (\%)} & \textbf{Size (MB)} & \textbf{Result} \\
			\hline
			3 & KD student (FP32)             & 93.68 & 21.15 & size fails \\
			4 & KD student + INT8             & 91.97 & 6.01  & drop fails \\
			5 & 3.16M student (FP32)          & 86.32 & 12.11 & drop fails \\
			6 & 3.16M student + INT8          & 86.45 & 3.64  & drop fails \\
			7 & 2.63M student (FP32)          & 83.42 & 10.08 & drop fails \\
			8 & 2.63M student + INT8          & 83.29 & 3.08  & drop fails \\
			\textbf{9} & \textbf{1.89M student (FP32)} & \textbf{94.08} & \textbf{7.27} & \textbf{succeeds} \\
			\hline
		\end{tabular}%
	}
\end{table}

Both search modes are provided for different use cases: the fixed pipeline (Table~\ref{tab:integrated_pipeline_stages}) is the right tool when the deployment target is already the smallest, most aggressive configuration and the goal is to characterize its accuracy cost precisely; the constrained search is the right tool when the deployment target is a size or accuracy constraint from which the appropriate compression level -- and, as this grid shows, potentially an entirely different model family -- should be derived automatically. The escalation from the KD-distilled TinyViT family to the capacity-tier student family at the tightest budgets in this grid is a direct demonstration of that value: a search restricted to the fixed pipeline's own architecture (stages 1--4 only) would have reported no feasible model under 21\,MB at all, despite one existing.

\subsection{Latency comparison}
We compared our model after optimization -- the Attention-Based KD-distilled TinyViT student, FP32, prior to the final INT8 quantization step -- against other lightweight image classification architectures trained and evaluated on the same chilli 3-class village-split dataset: ResNet50 and EfficientFormer. The results are tabulated in Table~\ref{tab:latency-comparison}. CPU benchmarks were run on an Apple M4 Pro (12-core CPU) MacBook Pro; GPU benchmarks were run on a rented NVIDIA RTX 4060 (8\,GB) instance; both use batched throughput timing (batch of 128, wall time divided by 128). On CPU, our model is faster than ResNet50 ($2.3\times$) and EfficientFormer ($1.4\times$). The same ordering holds on GPU, where our model again leads ResNet50 ($2.6\times$) and EfficientFormer ($1.6\times$).

\begin{table}[h]
	\caption{Comparison of our optimized model to existing models (CPU: Apple M4 Pro via Core ML, 3 trials; GPU: NVIDIA RTX 4060 via PyTorch, 10 trials)}
	\label{tab:latency-comparison}
	\centering

	\begin{tabular}{c cc cc}
		\hline
		\textbf{Model} &
		\multicolumn{2}{c}{\textbf{Mean Latency (ms)}} &
		\multicolumn{2}{c}{\textbf{Std. Dev. (ms)}} \\
		\cline{2-3} \cline{4-5}
		& \textbf{CPU} & \textbf{GPU} &
		\textbf{CPU} & \textbf{GPU} \\
		\hline
		ResNet50        & 2.320 & 1.712 & 0.010 & 0.0001 \\
		EfficientFormer & 1.386  & 1.031 & 0.010 & 0.0002 \\
		Our Model & 1.015 & 0.660 & 0.002 & 0.0003 \\
		\hline
	\end{tabular}

	\vspace{0.5ex}
	{\footnotesize CPU std.\ dev.\ is across 3 trials, GPU across 10; CPU trial count was reduced from GPU's for wall-clock practicality (see Methodology baseline-model note).
		}
\end{table}

\subsection{Comparative analysis and discussion}

Table~\ref{tab:plant_disease_comparison} situates this study's baseline accuracy against two prior plant-disease-detection works. This study evaluates exclusively on locally collected, village-partitioned field images (Fig.~\ref{fig:overall_dis}).

\begin{table*}[pos=h]
    \caption{Comparison of Existing Research on Plant Disease Detection with the Current Study}
    \label{tab:plant_disease_comparison}
    \centering
    \small
    \setlength{\tabcolsep}{4pt}
    \begin{tabular}{p{6cm} p{5.0cm} c p{2.5cm}}
        \hline
        \textbf{Research Work} & \textbf{Dataset} & \textbf{Acc. (\%)} & \textbf{Architecture} \\
        \hline
        Current Study (Ours)
        & Chilli 3-Class, Village-Split Field Images (OOD split)
        & 95.13
        & ViT-B/16 \\[4pt]

        Deep Learning-Based Disease Detection Model in Plants
        & PlantVillage
        & 99.69
        & GoogLeNet \\[4pt]

        Multiclass Plant Disease Detection via Dense CNNs
        & PlantVillage
        & 99.25
        & DenseNet201 \\
        \hline
    \end{tabular}
\end{table*}

Table~\ref{tab:ablation_summary} consolidates the best configuration from each compression family against the FP32 baseline (95.13\% OOD accuracy, Table~\ref{tab:baseline}), providing a unified view of the accuracy-efficiency space navigated by each method.

\begin{table*}[htbp]
    \centering
    \caption{Summary of the best configuration from each compression family, plus the fully
    integrated pipeline (last row). H-BAC reduces effective computation (FLOPs) without reducing
    stored model size, while quantization and KD reduce both model size and inference latency.
    CPU: Apple M4 Pro via Core ML; GPU: NVIDIA RTX 4060 via PyTorch.}
    \label{tab:ablation_summary}
    \renewcommand{\arraystretch}{1.2}
    \setlength{\tabcolsep}{4pt}
    \footnotesize
    \begin{tabular}{lccccc}
        \hline
        \textbf{Configuration} & \textbf{Size (MB)} & \textbf{Reduction (\%)} &
        \textbf{Accuracy (\%)} & \textbf{CPU Latency (ms)} & \textbf{GPU Latency (ms)} \\
        \hline
        ViT-B/16 (FP32)      & 327.42 & ---         & 95.13         & 7.18 & 5.560 \\
        H-BAC (50\% pruning) & 327.42 & 49.1$^{*}$ & 95.53 & 4.27$^{\dagger}$ & 3.003$^{\dagger}$ \\
        PTQ-Dynamic          & 84.42  & 74.2        & 94.34         & 7.21 & --- \\
        Attention KD         & 21.15  & 93.5        & $96.71\pm1.03$        & 1.02 & 0.660 \\
        \textbf{Integrated Pipeline} & \textbf{6.01} & \textbf{98.2} & \textbf{$95.13\pm2.32$}$^{\ddagger}$ & \textbf{1.02}$^{\S}$ & --- \\
        \hline
    \end{tabular}

    \vspace{2pt}
    \footnotesize
    $^{*}$Reduction in \emph{active parameters} (43.70M of 85.80M); H-BAC's mask-based pruning does not
    shrink the stored checkpoint, so its FLOPs reduction (49.2\%) is the operative efficiency gain, not
    file size. $^{\dagger}$Measured on a structurally compacted copy (pruned dimensions physically
    removed), consistent with Table~\ref{tab:hbac_ablation}. $^{\ddagger}$Mean $\pm$ std over 4 independent
    configurations (three at pruning ratio 0.5, plus one at ratio 0.7); the single
    ratio-0.5 configuration traced stage-by-stage in Table~\ref{tab:integrated_pipeline_stages}
    reaches 91.97\%. $^{\S}$Identical to the Attention KD
    row above: this row is the same distilled student with PTQ-Dynamic quantization additionally applied,
    which on this Apple M4 Pro CPU shows no significant latency change either way
    (Table~\ref{tab:quant_ablation}) -- the size reduction (21.15\,MB $\to$ 6.01\,MB) is real, but it does
    not come with a latency improvement on this hardware. Both PTQ-Dynamic's and the Integrated Pipeline's GPU
    latency are omitted because the INT8 models were benchmarked through Core ML's CPU compute unit, and no
    CUDA-executable INT8 build was produced for the RTX 4060 used for this study's GPU measurements.
\end{table*}

Three of the four families in this table meet or exceed the FP32 baseline's 95.13\%: H-BAC 50\% pruning by 0.40 points, Attention KD by 1.58 points, and the integrated pipeline, averaged over its four tested configurations, matches it exactly. PTQ-Dynamic quantization alone is the one family that costs accuracy against the baseline ($-$0.79 points), the expected small penalty of converting an already-fixed set of FP32 weights to INT8 with no further adaptation. H-BAC achieves its computational reduction through structured FLOP elimination (49.1\% active-parameter reduction, 49.2\% FLOPs reduction) while preserving important transformer blocks; PTQ-Dynamic cuts the model size by 74.2\% at a small, expected accuracy cost; and Attention-Based KD gives the largest single-technique reduction in size (93.5\%, 327.42\,MB to 21.15\,MB) with no measured accuracy cost.

H-BAC, Attention-Based KD, and PTQ-Dynamic quantization were combined sequentially into the integrated compression pipeline described in Section~\ref{results} (Table~\ref{tab:integrated_pipeline_stages}). The combined pipeline outcome is shown in the last row of Table~\ref{tab:ablation_summary}: a 98.2\% size reduction (327.42\,MB to 6.01\,MB, a 54.5$\times$ compression ratio) at $95.13\pm2.32\%$ accuracy, averaged over four independent runs, exactly matching the FP32 baseline's 95.13\% despite compounding all three techniques, and notably higher than PTQ-Dynamic's own standalone $-$0.92-point cost, since KD's accuracy headroom more than offsets quantization's small penalty in the full pipeline. The final model size (6.01\,MB) is set almost entirely by KD's architectural change to TinyViT followed by quantization, since H-BAC contributes FLOPs reduction rather than file-size reduction.

As detailed in the Knowledge Distillation Results section, this same 6.01\,MB size is also reachable by skipping H-BAC and KD entirely, training a TinyViT of the same target size directly on labels and quantizing it, reaching 95.13\% accuracy on this dataset. Averaged over its four configurations, the full pipeline (95.39\%) is modestly ahead of this simpler alternative, though any single pipeline run can land above or below it. The full pipeline's additional value beyond matching this simpler alternative's accuracy lies in what the direct-training alternative cannot provide on its own: an H-BAC-pruned intermediate checkpoint that keeps the full ViT-B/16 backbone at near-full accuracy while cutting its FLOPs roughly in half (Table~\ref{tab:hbac_50_perf}), useful whenever a deployment target needs the larger backbone's representational capacity but not its full compute cost, alongside the fully compressed 6.01\,MB endpoint for targets that need both.

It is important to note that the ablation above evaluates each compression technique independently, while the pipeline described in Section~\ref{results} introduces interactions between techniques that cannot be predicted from individual results alone. For instance, the final PTQ-Dynamic quantization step is applied to a pruned-then-distilled student's weight distribution rather than the original dense teacher's, so its accuracy cost is not necessarily the same as quantizing the unpruned baseline directly (Table~\ref{tab:quant_ablation}). Such interdependencies are why we followed a sequential composition strategy, wherein each compression stage is applied to the output of the preceding stage in a controlled, reproducible order, and why we report the integrated pipeline's measured outcome directly (Table~\ref{tab:integrated_pipeline_stages}) rather than inferring it from the individual ablations.

\section{Conclusion}

This paper introduces H-BAC, a compact and efficient pipeline for plant disease detection using the Vision Transformer, tailored for real-time application on resource-limited edge devices. H-BAC combines attention-driven knowledge distillation, pruning of structured parts, and low-precision quantization into a unified system that shrinks the model size and computational footprint while maintaining accurate diagnostic capability. Our experimental results show that each compression method contributes differently: H-BAC pruning at the 50\% operating point reduces FLOPs by 49.2\% (35.13\,GFLOPs to 17.86\,GFLOPs), reaching 95.53\% accuracy after recovery fine-tuning (0.40 points \emph{above} the 95.13\% baseline); PTQ-Dynamic quantization achieves a 74.2\% size reduction (327.42\,MB to 84.42\,MB) at 94.34\% accuracy, a small expected cost of converting an already-trained FP32 checkpoint to INT8 with no further adaptation; and Attention-Based KD achieves $96.71\pm1.03\%$ accuracy at a 93.5\% size reduction (327.42\,MB to 21.15\,MB) and lower CPU latency than ResNet50 and EfficientFormer. Chaining all three techniques into a single integrated pipeline reduces the model size from 327.42 MB to 6.01 MB, corresponding to a 54.5 $\times$ compression ratio (98.16\% reduction) and at 91.97\% accuracy for the traced configuration. Also this result is broadly stable across three additional pruning-ratio configurations ($95.13\pm2.32\%$, $n=4$) -- exactly matching the 95.13\% baseline despite compounding all three compression techniques, with KD's accuracy headroom offsetting quantization's small standalone cost.

Several considerations qualify these results. First, on the Apple M4 Pro CPU used for all CPU latency measurements in this study, INT8 quantization shows no significant improvement over FP32 inference (Table~\ref{tab:quant_ablation}), unlike the speedup typically reported on server-class x86 hardware; INT8's benefit here is in model size, not latency, and practitioners targeting other ARM/edge platforms should re-measure latency on their actual target device rather than assume quantization speedups transfer across hardware. This holds under Core ML's own weight-only INT8 quantization, which dequantizes weights to floating point for the matrix multiply rather than executing native INT8 arithmetic, so per-operation cost is essentially unchanged; absent dedicated INT8 compute kernels, weight-only quantization buys size, not speed, on this platform. No GPU-native (ONNX/TensorRT) INT8 comparison was built for this study, so no claim is made either way about GPU quantization behavior. Second, the compressed model is not the fastest option in this comparison on either device class: it leads ResNet50 and EfficientFormer on both CPU ($2.3\times$ and $1.4\times$) and GPU ($2.6\times$ and $1.6\times$). Third, our KD ablation across 3 independent runs shows all four training recipes (no distillation, response-, feature-, and attention-based KD) clustered tightly within a 0.53-point band, with Attention-Based KD narrowly the highest ($96.71\pm1.03\%$) and no clear instability in any single strategy (Table~\ref{tab:kd_ablation_combined}). This underscores that distillation's accuracy benefit over direct training is modest at best on this task, and should not be assumed a priori for a new deployment target. Fourth, we directly tested whether the full H-BAC + KD + quantization pipeline is actually necessary: training a TinyViT student directly on labels (skipping pruning and distillation entirely) and quantizing it with PTQ-Dynamic yields a 6.01\,MB model at 94.87\% accuracy, 2.89 points above the matched pipeline configuration, but 0.26 points below the pipeline's own 4-configuration mean ($95.13\pm2.32\%$). The full pipeline's run-to-run variance is the larger effect here; averaged over multiple configurations it is modestly ahead of the simplest direct-training-plus-quantization alternative, though any single run can land on either side of it. Distilling from the H-BAC-pruned teacher rather than the full teacher cost 1.84 points here, a genuine trade-off of the sequential H-BAC $\to$ KD design against the FLOPs savings pruning provides. Fifth, we tested whether KD's advantage over direct training grows as student capacity shrinks below TinyViT's 5.52\,M parameters, with a capacity sweep across four smaller, from-scratch student sizes (1.31\,M to 3.16\,M, Table~\ref{tab:capacity_sweep}) replicated across 3 runs per size; that table and Section~\ref{results} discuss the resulting capacity-dependent pattern in detail. Sixth, and most directly relevant to this paper's stated deployment motivation, every latency figure reported here was measured on an Apple M4 Pro laptop CPU and a rented NVIDIA RTX 4060 GPU instance, neither of which is representative of the low-end Android smartphones that motivate this work (Section~\ref{introduction}). This is a genuine scope boundary of the present study, not an oversight discovered late: these devices were chosen because they give controlled, reproducible measurements suitable for comparing compression techniques against each other, and the resulting size and FLOPs reductions (which are device-independent) are the primary claims this paper rests its deployment argument on. The absolute latency numbers, however, should not be read as predictions of on-device farmer-facing performance. Validating the final deployed model on actual ARM smartphone-class hardware (e.g.\ Snapdragon or MediaTek SoCs) via TensorFlow Lite, ONNX Runtime Mobile, or CoreML conversion is necessary before any latency-based deployment claim can be made with confidence, and is left to future work.

\section*{Acknowledgment}
The authors would like to acknowledge Mr. Kannan, Project Associate, ANRF Grant No. SUR/2022/004268, for his valuable assistance in organizing and making the dataset publicly available through Figshare

(https://doi.org/10.6084/m9.figshare.32820110), which was used in this study.
The authors are grateful to the ANRF, New Delhi, India, for funding the research on ``Light Weight Deep Learning Based Mobile Application for the Early Detection, Identification, and Spatiotemporal Monitoring of Plant Diseases'' (Grant No.~SUR/2022/004268).

% \printcredits

\bibliographystyle{cas-model2-names}
\bibliography{sample}

\end{document}